\documentclass{article}

\PassOptionsToPackage{numbers, compress}{natbib}
\usepackage{iclr2022_conference,times}
\iclrfinalcopy

\usepackage{amsmath,amsfonts,bm}

\def\eqref#1{equation~\ref{#1}}

\def\1{\bm{1}}

\DeclareMathAlphabet{\mathsfit}{\encodingdefault}{\sfdefault}{m}{sl}
\SetMathAlphabet{\mathsfit}{bold}{\encodingdefault}{\sfdefault}{bx}{n}

\def\sR{{\mathbb{R}}}

\usepackage[utf8]{inputenc} %
\usepackage[T1]{fontenc}    %
\usepackage[colorlinks=true,citecolor=blue]{hyperref}
\usepackage{url}            %
\usepackage{booktabs}       %
\usepackage{amsfonts}       %
\usepackage{nicefrac}       %
\usepackage{microtype}      %
\usepackage{xcolor}         %
\usepackage{graphicx}
\usepackage{float}
\usepackage{scalerel,amssymb}
\usepackage{algorithm}
\usepackage[noend]{algpseudocode}

\def\shownotes{1}  %
\ifnum\shownotes=1
\newcommand{\authnote}[2]{[#1: #2]}
\else
\newcommand{\authnote}[2]{}
\fi

\usepackage{multirow}
\usepackage{multicol}
\usepackage{tabularx}
\newcommand{\xhdr}[1]{\vspace{1.7mm}\noindent{{\bf #1.}}}
\usepackage{xspace}
\usepackage{caption}
\usepackage{graphicx}
\usepackage{booktabs}

\newcommand{\cut}[1]{}
\newcommand{\mb}{\mathbf}

\newcommand{\eg}{{e.g.}}
\newcommand{\ie}{\emph{i.e.}}
\newcommand{\ours}{MetaLink\xspace}
\newcommand{\name}{MetaLink\xspace}

\usepackage{titlesec}
\titlespacing\section{0pt}{4pt plus 2pt minus 1pt}{4pt plus 2pt minus 1pt}
\titlespacing\subsection{0pt}{3pt plus 2pt minus 1pt}{3pt plus 2pt minus 1pt}

\title{Relational Multi-Task Learning:\\
Modeling Relations between Data and Tasks}

\author{%
  Kaidi Cao\thanks{The two first authors made equal contributions.} \\
   \And
   Jiaxuan You\samethanks\\
   \And
   Jure Leskovec\\
}

\newcommand*\samethanks[1][\value{footnote}]{\footnotemark[#1]}

\begin{document}

\maketitle
\vspace{-3em}
 \begin{center}
 Department of Computer Science, Stanford University\\
 \texttt{\{kaidicao, jiaxuan, jure\}@cs.stanford.edu}
 \end{center}

\begin{abstract}

A key assumption in multi-task learning is that at the inference time the multi-task model only has access to a given data point but not to the data point's labels from other tasks.
This presents an opportunity to extend multi-task learning to utilize data point's labels from other auxiliary tasks, and this way improves performance on the new task.
Here we introduce a novel {\em relational multi-task learning} setting where we leverage data point labels from auxiliary tasks to make more accurate predictions on the new task. 
We develop \ours, where our key innovation is to build a knowledge graph that connects data points and tasks and thus allows us to leverage labels from auxiliary tasks. The knowledge graph consists of two types of nodes: (1) data nodes, where node features are data embeddings computed by the neural network, and (2) task nodes, with the last layer's weights for each task as node features. The edges in this knowledge graph capture data-task relationships, and the edge label captures the label of a data point on a particular task. Under \ours, we reformulate the new task as a link label prediction problem between a data node and a task node. The \ours framework provides flexibility to model knowledge transfer from auxiliary task labels to the task of interest.
We evaluate \ours on 6 benchmark datasets in both biochemical and vision domains. Experiments demonstrate that \ours can successfully utilize the relations among different tasks, outperforming the state-of-the-art methods under the proposed relational multi-task learning setting, with up to 27\% improvement in ROC AUC.

\end{abstract}

\section{Introduction}

The general idea of learning from multiple tasks has been explored under different settings, including multi-task learning~\citep{caruana1997multitask}, meta learning~\citep{finn2017model}, and few-shot learning~\citep{vinyals2016matching,cao2020concept,cao2020few}.
While these learning settings have inspired models that can utilize relationships among tasks~\citep{chen2019multi,zamir2018taskonomy,senermulti,lin2019pareto,ma2020efficient}, they are not able to capture the full complexity of real-world machine learning applications.
Concretely, when learning from multiple tasks, current approaches assume that the test data points have no access to the labels from other tasks when making predictions on a new task. However, this assumption oversimplifies potential useful knowledge in many real-world applications.

For example, multi-task learning studies simultaneously learning multiple predictive tasks to exploit commonalities between the tasks. At the test time the multi-task model predicts labels of a given data point for the tasks of interests, \eg, predicting whether the chemical compound $\mb{x}$ is non-toxic. At the same time, one may also have access to the data point’s labels for some other auxiliary tasks, \eg, the compound $\mb{x}$ has a positive result on two toxicology tests. Such auxiliary task labels could greatly help with improved predictions.

However, current deep learning architectures cannot model such knowledge transfer between auxiliary tasks/labels and the target tasks. %
Naively concatenating the known labels to the input features has its limitations, especially since such labels are sparsely available, and it is also unclear how to use the approach for new and unseen tasks. Another potential solution to model such flexible and conditional inference is through generative models~\citep{dempster1977maximum,koller2009probabilistic}. Although generative models are powerful, they are notoriously data-hungry, thus it is very difficult to construct and train a generative model for high dimensional data~\citep{turhan2018recent}. 

\begin{figure}[t]
    \centering
    \includegraphics[width=\textwidth]{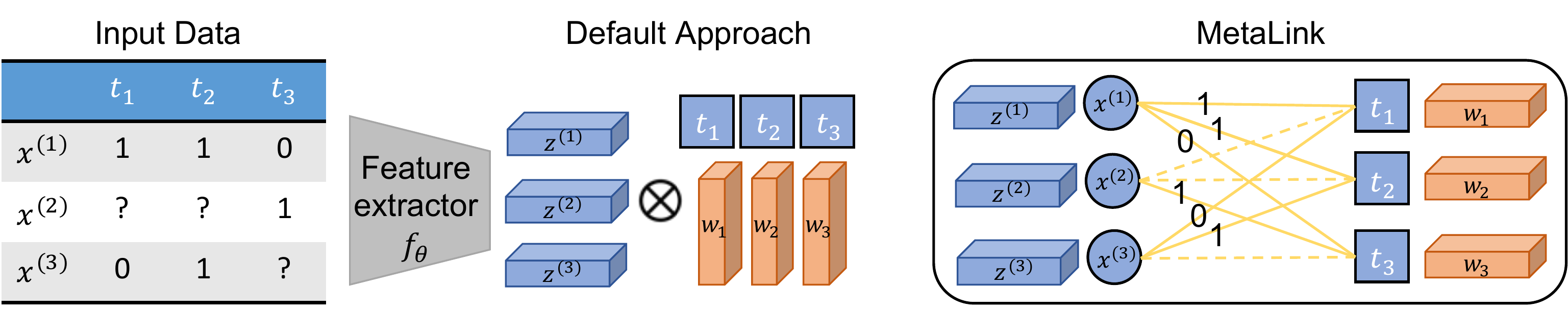}
    \caption{In the relational multi-task setting, the model learns to incorporate auxiliary knowledge in making predictions to achieve data efficiency. Concretely, given observations $\mb{x}^{(i)}$ and their labels $\{y^{(i)}_j\}$ (0/1 in this example) on subsets of tasks $\{t_j\}$, the goal is to build a model that can harness the auxiliary task labels $\{y^{(i)}_j\}$ and make predictions on a new task $t_n$. A standard approach is to build a multi-head deep neural network, with a prediction head for each individual task $t_j$. However, such approach cannot utilize auxiliary labels. In contrast, our proposed \name reinterprets the last layer's weights of each task as task nodes and creates a knowledge graph where data points and tasks are nodes and labeled edges provide information about labels of data points on tasks. Then, when predicting data point's label for a given task $t_j$, \name uses labels from other tasks to improve predictive performance.
    }
    \label{fig:teaser}
\end{figure}

Here we propose a new multi-task learning setting called \emph{relational multi-task learning}.
In our setting we distinguish between the target tasks (e.g., predicting molecule toxicity), which are tasks we aim to predict, and auxiliary tasks, which are tasks for which data point's labels are available at the inference time. Note that under our setting, each data point might have labels available for a different subset of auxiliary tasks. The goal then is to achieve strong predictive performance through leveraging labels of a given data point on some subset of auxiliary tasks.

To tackle the relational multi-task learning, we propose \ours\footnote{Source code is available at \url{https://github.com/snap-stanford/GraphGym}}, a general discriminative model that can explicitly incorporate the knowledge from auxiliary tasks.
Our key innovation is to build a knowledge graph that connects different tasks $t_j$ and data points $\mb{x}^{(i)}$ (Figure~\ref{fig:teaser}). The first step of our approach is to take input data points $\mb{x}^{(i)}$ and the feature extractor (\ie, neural network) $f_\theta$ to get its embedding $\mb{z}^{(i)}$. Then we build the knowledge graph which consists of two types of nodes: data nodes $\mb{x}^{(i)}$ and task nodes $t_j$. A data node $\mb{x}^{(i)}$ connects to a task node $t_j$ if data point $\mb{x}^{(i)}$ participates in task $t_j$ and the edge is annotated with the label $y^{(i)}_j$ of $\mb{x}^{(i)}$ on task $t_j$. 
We initialize data node features to be the last layer embedding $\mb{z}^{(i)}$ in the feature extractor $f_\theta$, and task node $t_j$ features are instantiated as the last layer's weights $\mb{w}_j$. 

Given our knowledge graph, we reformulate the multi-task learning problem as a link-label prediction problem between data nodes and task nodes. This means that at the inference time \ours is able to use all the information about a given data point $\mb{x}^{(i)}$ (including its labels on auxiliary tasks $\{ t_j \}$) to predict its label on a new task $t_n$.
We solve this link label prediction learning task via a Graph Neural Network (GNN)~\citep{hamilton2017inductive,he2019bipartite,xue2021multiplex}.
Unlike previous works, \eg, ML-GCN~\citep{chen2019multi}, that solely model relationships among tasks, \ours allows flexible and automatic modeling for data-task, data-data and task-task relationships.

We evaluate \ours on six benchmark datasets in both biochemical and vision domains under various settings. Empirical demonstrate that \ours can successfully utilize the relations among different tasks, outperforming the state-of-the-art methods under the proposed relational multi-task learning setting, with up to 27\% improvement in ROC AUC.

\section{Relational Multi-task Learning Settings}
\label{sec:setting}

\begin{figure}[tpb]
    \centering
    \vspace{-2mm}
    \includegraphics[width=\textwidth]{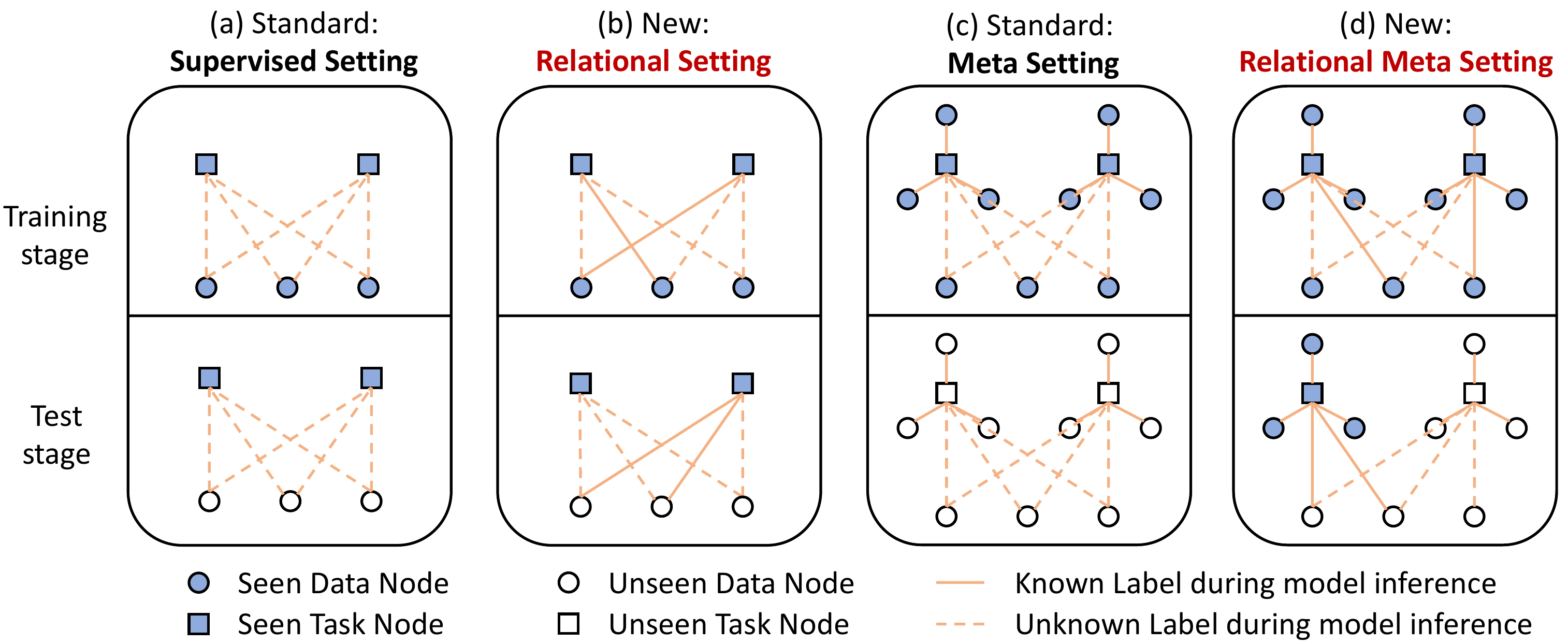}
    \caption{
    Our \ours framework allows for modeling four different multi-task learning settings:
    $\bigcirc$ represent data nodes and $\square$ represent task nodes. Blue represents the data/tasks seen in the training stage and white denotes the data/tasks seen only in the test stage. During model inference (for both the training and the test stage), the label of a data-task pair with solid line is known, while we want to predict labels of the data-task pairs with  dotted lines.}
    \label{fig:application}
    \vspace{-2mm}
\end{figure}

Here we first formally introduce the settings for relational multi-task learning.
Suppose we have $m$ machine learning tasks $\{t_j\}_{j\in T}$, where $T=\{ 1, 2, ..., m\}$ is integers between 1 and m.
We propose to categorize different settings in two dimensions: (1) whether the task is a relational task, \ie, if auxiliary task labels can be used at inference time; and, (2) whether the task is a meta task, \ie, if the task at test time has been seen at the training time.
Altogether, there are four possible task settings, which are illustrated in Figure \ref{fig:application} and below.

\xhdr{Standard supervised setting}
Let $\mb{x}^{(i)}$ denote the input and $y_j^{(i)}$ denote the corresponding label associated with task $t_j$, \ie, $y_j^{(i)} \sim t_j$.
Standard supervised multi-task learning can be represented as 
\begin{align*}
\text{\textbf{Train: }} \Big\{\mb{x}^{(i)} \rightarrow  \{ y_j^{(i)} \sim t_j \}_{j \in T} \Big\} \hspace{5mm}
\text{\textbf{Test: }} \Big\{\mb{x}^{(i)} \rightarrow \{ y_j^{(i)} \sim t_j \}_{j \in T}  \Big\}
\end{align*}
where $\rightarrow$ connects the input and the output. \emph{Training and test sets cover non-overlapping data points}. We use $y_j^{(i)}$ to concisely represent $y_j^{(i)} \sim t_j $ later on.

\xhdr{Relational setting}
In relational setting, in addition to input $\mb{x}^{(i)}$, we assume we also have access to auxiliary task labels when making predictions. 
$T_{\text{aux}}$ and $T_{\text{test}}$ are partitions of integers ($T$) that relates to subsets of tasks.
Specifically, $T_{\text{aux}}$ refers to the indices of tasks that input $\mb{x}$ has access to, and $T_{\text{test}}$ are the indices of tasks that we wish to predict; these two sets are non-overlapping, \ie, $T_\text{aux} \cap T_\text{test} =\emptyset$, and input-dependent, \ie, the partition $T_{\text{test}}^{(1)}$ and $T_{\text{test}}^{(2)}$ can be different. The input-output pairs are now in the form of 
\begin{align*}
\text{\textbf{Train: }} \Big\{(\mb{x}^{(i)},  \{ y_j^{(i)}  \}_{j \in T^{(i)}_{\text{aux}}}  ) &\rightarrow  \{ y_j^{(i)}  \}_{j \in T^{(i)}_{\text{test}}}\Big\} \hspace{5mm} \text{\textbf{Test: }}  \Big\{(\mb{x}^{(i)},  \{ y_j^{(i)}  \}_{j \in T^{(i)}_{\text{aux}}} ) &\rightarrow  \{ y_j^{(i)}  \}_{j \in T^{(i)}_{\text{test}}}\Big\} 
\end{align*}
\xhdr{Meta setting} %
In the meta setting, we want to learn to predict unseen tasks at the test time.
Formally, let $T_{\text{s}}$, $T_{\text{u}}$ denote the set of partitions for seen tasks (used at training time) and unseen tasks (used at test time), where $T_{\text{s}} \cap T_{\text{u}} = \emptyset$.
Here, we have access to a batch of samples with labels as the support set $S$ and the objective is to correctly predict samples in the query set $Q$.
\begin{align*}
\text{\textbf{Train: }given } S = \Big\{ (\mb{x}^{(i)},  \{ y_j^{(i)}  \}_{j \in T_{\text{s}}} ) \Big\}, \text{ predict }Q = \Big\{\mb{x}^{(i)} \rightarrow   \{ y_j^{(i)}  \}_{j \in T_{\text{s}}} \Big\} \\
\text{\textbf{Test: }given }S = \Big\{ (\mb{x}^{(i)},  \{ y_j^{(i)}  \}_{j \in T_{\text{u}}} ) \Big\}, \text{ predict }Q = \Big\{\mb{x}^{(i)} \rightarrow   \{ y_j^{(i)}  \}_{j \in T_{\text{u}}}\Big\} 
\end{align*}
\xhdr{Relational meta setting}
Relational meta setting combines the features of relational setting and meta setting.
Similar to the meta setting, we aim to predict unseen tasks  $T_{\text{u}}$ at test time; meanwhile, similar to the relational setting, we also assume having labels on a limited number of auxiliary tasks $T_{\text{aux}} \subseteq T_{\text{s}}$ to harness. Formally, we have a support set and a query set in the form of
\begin{align*}
\text{\textbf{Train: }given } S = \Big\{ (\mb{x}^{(i)},  \{ y_j^{(i)}  \}_{j \in T^{(i)}_{\text{s}}} )\Big\}, \text{ predict }Q &= \Big\{(\mb{x}^{(i)},  \{ y_j^{(i)}  \}_{j \in T^{(i)}_{\text{aux}}} )  \rightarrow   \{ y_j^{(i)}  \}_{j \in T^{(i)}_{\text{s}}\setminus T^{(i)}_{\text{aux}}}  \Big\} \\
\text{\textbf{Test: }given }  S = \Big\{ (\mb{x}^{(i)},  \{ y_j^{(i)}  \}_{j \in T^{(i)}_{\text{u}}} ) \Big\}, \text{ predict }Q &= \Big\{(\mb{x}^{(i)},  \{ y_j^{(i)}  \}_{j \in T^{(i)}_{\text{aux}}} )  \rightarrow   \{ y_j^{(i)}  \}_{j \in T^{(i)}_{\text{u}}}\Big\} 
\end{align*}

\section{\ours Framework}
\label{sec:framework}
Next, we describe our \ours framework, which allows us to formulate the above multi-task learning settings in a single framework. In particular, \ours formulates them as a link label prediction task on a heterogeneous knowledge graph, this way, \ours can harness the relational information about data and tasks. 

\subsection{Build a Knowledge Graph on Task Heads}

We first recap the general formulation of a neural network.
Given data points $\{ \mb{x}^{(i)}\}_{i=1}^n$ and labels $\{ \{ y^{(i)}_j \}_{j \in T} \}_{i=1}^n$, a deep learning model can be formulated as a parameterized embedding function $f_{\theta}$ (which is deep) and a task head $f_{\mb{w}}$, consisting of only a single weight matrix $\mb{w}$.
$f_{\theta}$ maps a data point $\mb{x}^{(i)}$ to a vector embedding space, $f_{\theta}(\mb{x}^{(i)}) = \mb{z}^{(i)} \in \sR^D$.
$f_{\mb{w}}$ then maps an embedding $\mb{z}^{(i)}$ to prediction $\hat{y}^{(i)}  \in \sR$\footnote{Without loss of generality, we assume a task is a binary classification or 1-D regression}, $f_{\mb{w}}(\mb{z}^{(i)}) = \mb{w}^T \mb{z}^{(i)} =  \hat{y}^{(i)}$. When a task head involves multi-layer transformation, we have $f_{\mb{w}}(\mb{z}^{(i)}) = \mb{w}^T g(\mb{z}^{(i)}) = \hat{y}^{(i)}$, where $g(\cdot)$ can be an arbitrary function.
In multi-task learning settings, people usually assign multiple task heads to a neural network. Suppose we have $m$ tasks $\{t_j\}_{j=1}^m$, then there are $m$ task heads such that $\hat{y}^{(i)}_j = f_{\mb{w}_j}(\mb{z}^{(i)})$. 

Here, our observation is that the weights in task heads $\mb{w}_1,\dots,\mb{w}_j$, and the data embeddings $\mb{z}^{(i)}$ play symmetric roles in a multi-task prediction task (due to the dot product).
Therefore, instead of viewing weights $\mb{w}_1,\dots,\mb{w}_j$ as parameters in a neural network, we propose to represent $\mb{w}_1,\dots,\mb{w}_j$ as another type of input that supports the prediction.
Essentially, we reformulate a task head from $\hat{y}^{(i)}_j = f_{\mb{w}_j}(\mb{z}^{(i)})$ to $\hat{y}^{(i)}_j = f_{\phi}(\mb{w}_j, \mb{z}^{(i)})$.
This new perspective, where both task weights $\mb{w}_j$ and embedding $\mb{z}^{(i)}$ are viewed as input, enables us to build a more sophisticated predictive model $f_{\phi}$. $f_{\phi}$ in the sense that it contains two main steps, \ie, $\text{GraphConv}( \cdot )$ and $\text{EdgePred}( \cdot )$. In general, $\text{EdgePred}( \cdot )$ has a similar model complexity as $f_{\mb{w}_j}$, whereas $\text{GraphConv}( \cdot )$ provides additional expressiveness.

In \ours, we propose to build a \emph{knowledge graph} over task weights $\mb{w}_j$ and embedding $\mb{z}^{(i)}$.
By building this knowledge graph, we can succinctly represent relationships between data points and tasks, as well as different multi-task learning settings.
Concretely, a knowledge graph helps us easily express any data-task relationship (\eg, a data point has a label on a given task), data-data relationship (\eg, two data points are similar or not), or task-task relationship (\eg, hierarchy of different tasks).
Moreover, a knowledge graph greatly simplifies the representation of all the multi-task learning settings that we outlined in Section \ref{sec:setting}; in fact, all the settings can be viewed as link label prediction tasks where different portions of the knowledge graph can be constructed, as illustrated in Figure \ref{fig:application}.

We define the knowledge graph as $G = \{ V, E \}$, where $V$ is the node set and $E \subseteq V \times V$ is the edge set.
We define two types of nodes, data nodes $V_d=\{\mb{x}^{(1)},\dots,\mb{x}^{(n)}\}$ and task nodes $V_t=\{t_1,\dots,t_m\}$.
We can then define edges between data and task nodes $E_{dt} \subseteq V_d \times V_t$, within data nodes $E_{dd} \subseteq V_d \times V_d$, and within task nodes $E_{tt} \subseteq V_t \times V_t$.
\ours framework can work with all three types of edges; however, since most benchmark datasets do not have information on data-data or task-task relationship, we focus on data-task relationship $E_{dt}$ in the remaining discussions. 
Specifically, we define $E_{dt}$ based on task labels, $ E_{dt} = \{(\mb{x}^{(i)}, t_j) \sim  y^{(i)}_{j \in T_{\text{aux}}}\}$, \ie, we connect a data node $\mb{x}^{(i)}$ with a task node $t_j$ if label $y^{(i)}_j$ exists.

\subsection{Learn from the Knowledge Graph via a Heterogeneous GNN}

Given the constructed graph, we then discuss how \ours learns from the built knowledge graph.

\xhdr{Initialize node/edge features}
First, we initialize features for the knowledge graph. Concretely, we initialize data node features to be the data embeddings $\mb{z}^{(i)}$ computed from the feature extractor $f_\theta(\cdot)$, $\mb{h}_i^{(0)}=\mb{z}^{(i)}$. 
We initialize the seen task node embeddings using the weights in task heads $\mb{w}_j$, $\mb{h}_j^{(0)}=\mb{w}_j$. In meta settings where unseen task nodes appear during test time, we will initialize those nodes by a constant vector of $\textbf{1}$; such initialization ensures that \ours can generalize to unseen tasks, since the node feature construction process is inductive.

\xhdr{Predict via a heterogeneous GNN}
We implement the predictive model $f_{\phi}$ over data nodes and task nodes as a GNN.
The goal of GNN is to learn expressive node embeddings $\mb{h}_v$ based on an iterative aggregation of local network neighborhoods. The $l$-th iteration of $\text{GraphConv}( \cdot )$, or the $l$-th layer, can be written as:
\begin{equation}
    \label{eq:gnn}
    \begin{aligned}
    & \mb{h}_v^{(l)} = \textsc{Agg}^{(l)}\Big(\{\textsc{Msg}^{(l)}(\mb{h}_u^{(l-1)}), u \in \mathcal{N}_G(v)\}, \mb{h}_v^{(l)}\Big)
    \end{aligned}
\end{equation}
where $\mb{h}_v^{(l)}$ is the node embedding after $l$ iterations, $\mb{h}_v^{(0)}$ have been initialized as explained above, and $\mathcal{N}_G(v)$ denotes the direct neighbors of $v$. $\textsc{Agg}$ is the acronym for aggregation function and $\textsc{Msg}$ is the acronym for message function.
We perform $L$ GNN layers on top of the knowledge graph $G$ that we have built. After updating data and task node embeddings, we can make predictions on the given task through $\text{EdgePred}( \cdot )$ in the form of $\hat{y}_j^{(i)} = \textsc{MLP} \Big( \textsc{Concat} (\mb{h}_i^{(L)}, \mb{h}_j^{(L)}) \Big)$.

In general, \ours should work with any GNN architecture that follows the definition in Equation \ref{eq:gnn}. We use the GraphSAGE layer \citep{hamilton2017inductive} in \ours ($\mb{W}^{(l)}$, $\mb{U}^{(l)}$ are trainable):
\begin{equation}
    \label{eq:sage}
    \mb{h}_v^{(l)} = \mb{U}^{(l)}\textsc{Concat}\Big(\textsc{Mean}\Big(\{\textsc{ReLU}(\mb{W}^{(l)}\mb{h}_u^{(l-1)}),u \in \mathcal{N}(v)\}\Big), \mb{h}_v^{(l-1)}\Big)
\end{equation}
Next, we discuss the special GNN design used in \ours which has shown to be successful.

\xhdr{Special GNN designs in \ours}
We make three extensions in the formulation in Equation \ref{eq:gnn}. 
First, since there are two types of nodes in our knowledge graph, we define different message passing functions for different message types, \ie, the message from data nodes to task nodes, and the message from task nodes to data nodes. 
Second, we include edge features in the message computation. This is especially important for our formulation, since the task label values $y_j^{(i)}$ are included as edge features, and should be considered during GNN message passing. 
Concretely, we extend Equation \ref{eq:sage} into:
\begin{equation}
    \mb{h}_v^{(l)} = \mb{U}^{(l)}\textsc{Concat}\Big(\textsc{Mean}\Big(\{\textsc{ReLU}(\mb{W}_{\mathbf{1}[v\in V_d, u\in V_t]}^{(l)}\mb{h}_u^{(l-1)}+\mb{O}^{(l)}y_v^{(u)}),u \in \mathcal{N}(v)\}\Big), \mb{h}_v^{(l-1)}\Big)
\end{equation}
where $\mathbf{1}[v\in V_d, u\in V_t]$ indicates the message type (whether from task to data, or data to task), and $\mb{O}^{(l)}$ is an additional trainable weight that allows task label $y_v^{(u)}$ participates in message passing.
Finally, we let each GNN layer make a prediction and sum them up as the final prediction; this way, the final prediction is obtained from mixed information from different hops of node neighbors. We observe this multi-layer ensemble technique can help \ours make robust predictions.

\xhdr{\ours for relational meta setting}
Here we provide a detailed description on how to apply \ours for the relational meta setting in Algorithm~\ref{alg:metalink}.
At training time, since most of the existing multi-label datasets are not designed for meta setting, we manually simulate such setting by sampling a mini batch with support $S = \Big\{ (\mb{x}^{(i)},  \{ y_j^{(i)}  \}_{j \in T^{(i)}_{\text{s}}} ) \Big\}$ and query set $\Big\{(\mb{x}^{(i)},  \{ y_j^{(i)}  \}_{j \in T^{(i)}_{\text{aux}}} )  \rightarrow   \{ y_j^{(i)}  \}_{j \in T^{(i)}_{\text{s}}\setminus T^{(i)}_{\text{aux}}} \Big\} $. We make sure the sampled meta tasks $T^{(i)}_s$ and auxiliary knowledge tasks $T^{(i)}_{\text{aux}}$ have no intersection. We first use the feature extractor $f_\theta$ to get data embeddings and use the embeddings to initialize data nodes. To initialize task nodes, we either: (1) use $\textbf{1}$ to initialize if the task is a meta task; Or, (2) use trained weights otherwise. We construct the edge set $E$ by connecting each data-task pair based on $\Big\{ \{ y_j^{(i)}  \}_{j \in T^{(i)}_{\text{s}}} \Big\}$ in the support set and $\Big\{ \{ y_j^{(i)}  \}_{j \in T^{(i)}_{\text{aux}}} \Big\}$ (we eliminate the relations we want to predict) in query set. Now that we have all the components of the knowledge graph, we can apply the predictive graph model $f_\phi$ to learn expressive data and task node embeddings and make predictions.

At test time, we use the same pipeline as in training to construct the knowledge graph and run inference.
Please refer to Appendix~\ref{sec:implementation} for the pipelines of relational or meta settings.

\begin{algorithm}[h]
\caption{\ours{} Training in Relational Meta Setting}
\label{alg:metalink}
\begin{algorithmic}[1]
\Require Dataset $\mathcal{D}_\text{train} = \{(\mb{x},y)\}$. A parameterized embedding function $f_\theta$. Last layer weights for each task $
\mb{w}_j$. A parameterized heterogeneous GNN $f_\phi$. Number of GNN layers $L$. 

\For {each iteration }
    
    \State $S, Q \leftarrow \text{SampleMiniBatch}(\mathcal{D}_\text{train}) $ \Comment{Simulate meta setting in training}
    \State $ \{ \mb{z} \}  \leftarrow f_\theta ( \mb{x} ) \text{ for } \mb{x} \in (S, Q)$
    \State $V_d^{(0)} = \{ \mb{h}_i^{(0)} \leftarrow \mb{z} \text{ for } \mb{z} \in \{ \mb{z} \} \}$ \Comment{Initialize data nodes}
    \State $V_t^{(0)} = \{ \mb{h}_j^{(0)} \leftarrow \textbf{1} \text{ if meta else } \mb{w}_j \text{ for each } \mb{w}_j \}$ \Comment{Initialize task nodes}
    \State $E = \{ \mb{e}_{ij} \leftarrow (\mb{x}^{(i)}, t_j) \text{ for }  y^{(i)}_j \in (S,Q) \} $ \Comment{Initialize edges}
    
    \For {$l=1$ to $L$ }
        \State $ V_d^{(l)}, V_t^{(l)} \leftarrow  \text{GraphConv} (V_d^{(l-1)}, V_t^{(l-1)}, E)$ with $f_\phi$
    \EndFor
    \State $\text{logits} \leftarrow \text{EdgePred}( V_d^{(L)}, V_t^{(L)})$ with $f_\phi$
    \State $\text{Backward}\left( \text{Criterion}(\text{logits}, \{  \{ y_j^{(i)} \}_{j \in T^{(i)}_{\text{s}}\setminus T^{(i)}_{\text{aux}}} \} \in Q) \right)$
\EndFor

\end{algorithmic}
\end{algorithm}

\section{Experiments}
\label{sec:experiments}

Here we experimentally show that our proposed \ours{} flexibly handles different settings and leverages knowledge from auxiliary tasks.
We first evaluate our algorithms on Tox21~\citep{huang2016tox21challenge}, Sider~\citep{kuhn2016sider}, ToxCast~\citep{richard2016toxcast}, and MS-COCO~\citep{lin2014microsoft} datasets with various controllable settings on relational multi-task learning.
To additionally demonstrate the advantage of \ours, we also include experiments on a well-studied task: few-shot learning.
Our core algorithm is developed using PyTorch~\citep{paszke2017automatic}. We use one NVIDIA RTX 8000 GPU for each experiment and the most time-consuming one (MS-COCO) takes less than 24 hours.
Please refer to Appendix~\ref{sec:implementation} for additional low-level implementation details. 

\subsection{Evaluating \ours on Relational Multi-task Settings}

\xhdr{Datasets} We simulate four relational multi-task settings (Figure~\ref{fig:application}) using four widely-used multi-label datasets. Tox21~\citep{huang2016tox21challenge} contains 12 different toxicological experiments for each sample with binary labels (active/inactive). %
Sider~\citep{kuhn2016sider} is a database of marketed drugs and adverse drug reactions (ADR), grouped into 27
tasks. ToxCast~\citep{richard2016toxcast} contains about 8K pairs of molecular graphs and corresponding 617-dimensional binary vectors that represent different
experimental results.  Microsoft COCO (Common Objects in Context)~\citep{lin2014microsoft} is originally a large-scale object detection, segmentation dataset. By counting whether each type of object exists in a scene as a single task, it also serves as the default large-scale dataset for benchmarking multi-label classification in vision. There are 80 binary classification tasks with an average of 2.9 positive labels per image.

\xhdr{Experimental setting} We evaluate \ours{} on all four settings described in Figure~\ref{fig:application}, and we summarize the detailed configurations as follows. \textsc{Standard setting}: standard supervised learning. In order to have a fair comparison, we evaluate the same set of tasks with unknown labels as we sampled in relational settings. \textsc{Relational setting}: we assume each example has access to labels of 20\% of all tasks in each dataset. We evaluate on the rest of the tasks with unknown labels. \textsc{Meta setting}: we hold out 20\% of the tasks at the training time. We only evaluate on the held-out tasks at the test time. We use 256-shot setting, meaning at test time, we use 256 data points as a support set to initialize the prototypes of unseen tasks. The reason why the number of shots is much larger than what is commonly used in few-shot learning (1-shot, 5-shot) is that positive labels are sometimes sparse in certain tasks. \textsc{Relational meta setting}: we hold out the same 20\% of the tasks at the training time as in the meta setting. At test time, we assume each unseen task has access to 20\% of the labels of seen tasks. We will release our splits of meta setting to promote reproducibility.

\xhdr{Baselines} Though the main motivation of our work is to utilize auxiliary task labels, we still include a few baselines under the standard supervised setting in order to benchmark state-of-the-art results. The simplest one is (1) Empirical risk minimization (ERM): we train the network with cross-entropy loss under the standard supervised setting; (2) Various commonly used graph neural network architectures designed for molecules, MPNN \citep{gilmer2017neural}, DMPNN \citep{yang2019analyzing}, MGCN \citep{lu2019molecular}, AttentiveFP \citep{xiong2019pushing}; (3) GROVER \citep{rong2020self} integrates Message Passing Networks into the Transformer-style architecture. By leverage pretraining, it achieves state-of-the-art results on the aforementioned molecule datasets; (4) Baseline++~\citep{chen2018closer}: since there is no prior work on addressing meta setting for multi-task learning, we adapt Baseline++ from few-shot learning to this setting. We first train a feature extractor on the training set. At test time we use the support set to train a linear classification layer. %

\xhdr{Results on biochemical datasets}
Table~\ref{tab:graph_property} summarizes the results on Tox21, Sider, and ToxCast datasets. We first clarify that when performing \ours{} under the standard setting, where there is no auxiliary task label to leverage, the knowledge graph we built degenerates to a trivial set since the edge set is empty. It has a negligible difference compared with training a vanilla network with cross-entropy loss. Thus, it is reasonable that \ours{} under the standard setting has similar performance with the baselines. GROVER has better performance because it leverages large-scale pretraining. In terms of comparing relational setting with standard setting, we conclude that the proposed \ours{} successfully leverages the auxiliary knowledge for each sample. Notably, with auxiliary labels from 20\% of the tasks, \ours{} surpassed GROVER with large-scale pretraining by a large margin. In addition, we observe similar improvements between meta and relational meta settings. The empirical results for meta or relational meta setting are very close to the performance in standard setting, proving the potential of data-efficient research in biochemical domains. 

\xhdr{Results on MS-COCO}
Table~\ref{tab:coco} summarizes the results on MS-COCO. Note that as also mentioned in the settings, we need a large support set to ensure each task in the meta testing stage has at least one positive label. This blocks the feasibility of using a large model and input size as it is done in common benchmarks. Thus, we use ResNet-50 with an input size of 224 for all the experiments.
We observe consistent improvements for both relational setting and relational meta setting. \ours{} also outperforms Baseline++ in the meta setting. Note that since label distribution for most of the tasks is highly skewed, mAP degrades more in the meta setting than for the biochemical datasets. 

\begin{table}[t]
\caption{Results of different multi-task learning settings on graph classification tasks, measured in ROC AUC. \ours can successfully utilize the relations among different tasks, outperforming the state-of-the-art method GROVER~\citep{rong2020self} under the relational setting.
}
\centering
\begin{footnotesize}

\resizebox{\textwidth}{!}{
\begin{tabular}{cc|cccc}
\toprule

\texttt{Method} & \texttt{Setting} & \texttt{Tox21} (12 tasks) & \texttt{Sider} (27 tasks) & \texttt{ToxCast} (617 tasks)  \\ \midrule

MPNN \citep{gilmer2017neural} & \multirow{6}{*}{Standard} & 80.8$\pm$2.4 & 59.5$\pm$3.0 & 69.1$\pm$1.3  \\
DMPNN \citep{yang2019analyzing} &  & 82.6$\pm$2.3 & 63.2$\pm$2.3 & 71.8$\pm$1.1 \\ 
MGCN \citep{lu2019molecular} &  & 70.7$\pm$1.6 & 55.2$\pm$1.8 & 66.3$\pm$0.9 \\ 
AttentiveFP \citep{xiong2019pushing} &  & 80.7$\pm$2.0 & 60.5$\pm$6.0 & 57.9$\pm$1.0 \\  
GROVER(48M) \citep{rong2020self} &  & 81.9$\pm$2.0 & 65.6$\pm$0.6 & 72.3$\pm$1.0 \\  
GROVER(100M) \citep{rong2020self} &  & \emph{83.1}$\pm$2.5 & \emph{65.8}$\pm$2.3 & \emph{73.7}$\pm$1.0 \\ \midrule 
\multirow{7}{*}{\textbf{\ours{}}} & Standard & 82.3$\pm$2.2 \textsubscript{(KG layer = 1)} & 60.9$\pm$2.4 \textsubscript{(KG layer = 5)} & 69.3$\pm$1.6 \textsubscript{(KG layer = 3)} \\ 
& \textbf{Relational} & \textbf{83.7}$\pm$\textbf{1.9} \textsubscript{(KG layer = 1)} & \textbf{76.8}$\pm$\textbf{3.0} \textsubscript{(KG layer = 2)} & \textbf{79.4}$\pm$\textbf{1.0} \textsubscript{(KG layer = 4)} \\
\cmidrule{2-5}
& Meta & 77.5$\pm$2.1 \textsubscript{(KG layer = 2)} & 57.9$\pm$5.0 \textsubscript{(KG layer = 2)} & 71.3$\pm$2.2 \textsubscript{(KG layer = 2)} \\ 
& \textbf{Relational}+Meta & \textbf{79.2}$\pm$\textbf{2.9} \textsubscript{(KG layer = 2)} & \textbf{65.4}$\pm$\textbf{4.3} \textsubscript{(KG layer = 5)} & \textbf{84.3}$\pm$\textbf{1.2} \textsubscript{(KG layer = 5)} \\

\bottomrule
\end{tabular}
}

\label{tab:graph_property}
\end{footnotesize}
\end{table}

\begin{table}[htpb]
\centering
\fontsize{9}{8}\selectfont
\caption{Results on MS-COCO (80 tasks) dataset. We report the average accuracy and standard deviation over 5 runs on the validation set. We use ResNet-50 and our input size is 224. \ours achieves the best performance in relational settings through harnessing the auxiliary labels efficiently.}
\label{tab:coco}
\begin{tabular}{cc|c}
\toprule
\texttt{Method}           & \texttt{Setting}    & \texttt{mAP}   \\  \midrule
ML-GCN~\citep{chen2019multi}\textsuperscript{1} & Standard & 69.15 $\pm$ 0.19  \\ 
ERM & Standard & 71.22 $\pm$ 0.15  \\ 
Baseline++~\citep{chen2018closer} & Meta & 30.46 $\pm$ 0.69  \\ \midrule
\multirow{6}{*}{\textbf{\ours{}}} & Standard & 71.58 $\pm$ 0.16 \\
 & \textbf{Relational} & \textbf{75.36} $\pm$ \textbf{0.16} \\
\cmidrule{2-3}
 & Meta & 41.75 $\pm$ 0.92 \\
 & \textbf{Relational}+Meta & \textbf{49.73} $\pm$ \textbf{0.88} \\
\bottomrule
\end{tabular}
\end{table}
\footnotetext[1]{\url{https://github.com/Megvii-Nanjing/ML-GCN}}

\subsection{Ablation Studies}

\textbf{Does \ours truly learn to utilize correlations between tasks?} To better understand the improvement of our algorithm, we first plot the Pearson correlation heat map on Sider (Figure~\ref{figu:corr}). We then find the top 3 tasks with the highest and lowest average correlation with respect to the rest of the tasks. We report the performance of \ours{} on these two subsets of tasks, respectively (Table~\ref{tab:corr}). We observe that \ours demonstrates larger improvement on tasks with higher correlations. This experiment verifies that \ours learns to utilize correlations between tasks as expected.

\begin{table}[t]
    \begin{minipage}{0.45\linewidth}
		\centering
		\captionof{figure}{Pearson correlation heat map on 27 tasks of Sider dataset.}
		\label{figu:corr}
		\includegraphics[width=0.8\textwidth]{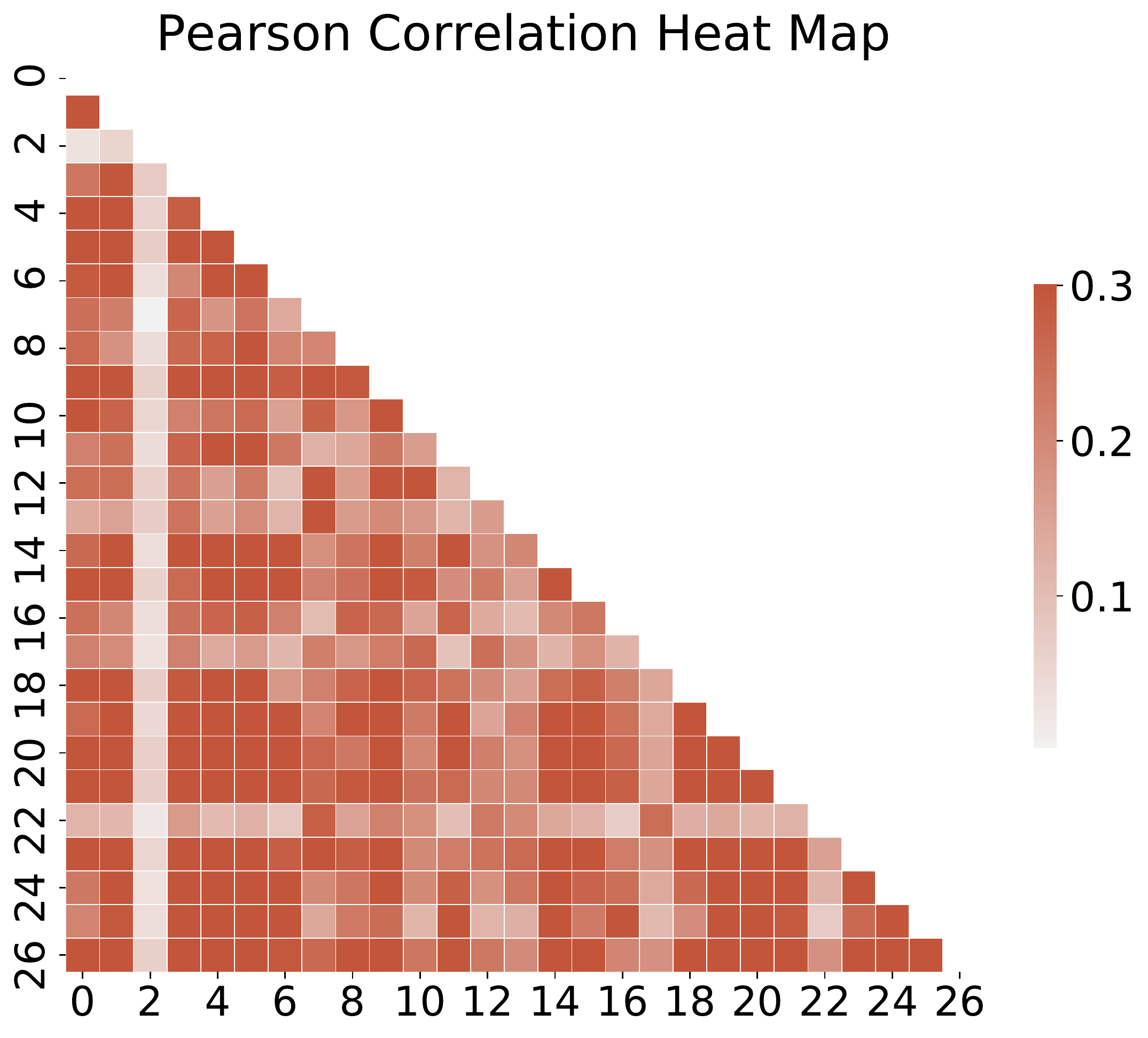}
	\end{minipage}
    \hfill
	\begin{minipage}{0.5\linewidth}
		\caption{ROC AOC on 3 tasks with the lowest average Pearson correlation to other tasks (tasks 2, 17, 22), as well as 3 tasks with the highest average Pearson correlation to other tasks (tasks 5, 20, 21) on the Sider dataset. We observe \ours gains more improvement on the tasks with high Pearson correlation.}
		\label{tab:corr}
		\centering
		\resizebox{\textwidth}{!}{%
		\begin{tabular}[width=\linewidth]{@{}ccc@{}}
    \toprule
    \texttt{Setting} & \texttt{Low Corr Tasks} & \texttt{High Corr Tasks} \\ 
    \midrule
    Standard & 57.5 $\pm$ 8.4 & 61.0 $\pm$ 4.1 \\
    \textbf{Relational} & \textbf{70.1 $\pm$ 7.8}  & \textbf{80.6 $\pm$ 3.3} \\
     Improvement  & 21.9\% & 32.1\% \\ 
    \bottomrule
    \end{tabular}}
	\end{minipage}
\end{table}

\begin{figure}[h]
    \centering
    \includegraphics[width=\textwidth]{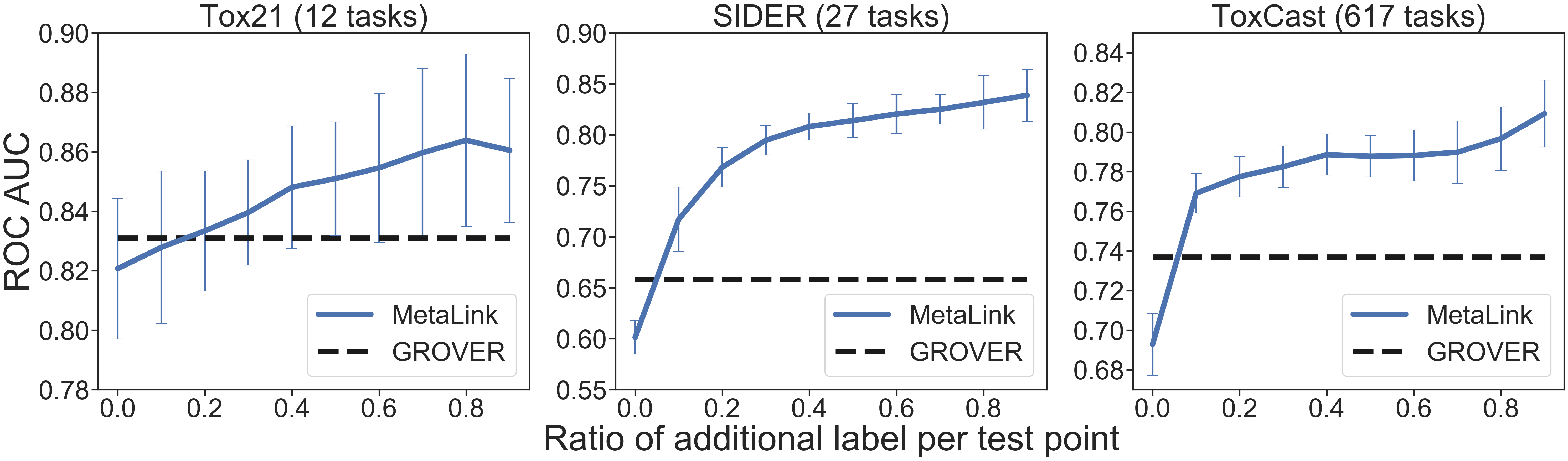}
    \caption{We vary the ratio of auxiliary labels per test point and plot the ROC AUC with the error bar. We also plot the state-of-the-art method (GROVER) as the dash line which cannot utilize additional auxiliary task labels. The performance of \ours consistently improves as more tasks are utilized. In Sider, MetaLink can outperform GROVER by up to 27\% when 90\% auxiliary labels are provided.}
    \label{fig:relation_ratio}
\end{figure}

\textbf{How does \ours perform with varying ratio of auxiliary task labels?} 
We vary the ratio of additional labels per test point and report the results in Figure~\ref{fig:relation_ratio}. We observe consistent improvement as we gradually increase the number of auxiliary labels in each dataset. This experiment demonstrates that \ours can successfully utilize marginal information whenever auxiliary task labels are added to the dataset. The improvement is usually prominent for the first few added labels.

\subsection{Evaluating \ours on Standard Few-shot Learning Datasets}

Since the relational multi-task setting is novel, there is a very small number of baselines that we can compare against. The major aim of this section is to show the advantage of \ours{} in a well-studied problem: few-shot learning. Note that there is a slight difference between the meta setting above and few-shot learning setting here. For few-shot learning, for all the link label prediction tasks related to an input, there will be only one positive link. This inductive bias could be easily incorporated by modeling all the link label predictions together using cross-entropy loss.

\xhdr{Experimental setting} We evaluate performance on two standard benchmarks: mini-ImageNet~\citep{vinyals2016matching} and tiered-ImageNet~\citep{ren2018meta}. We compare against MatchNet~\citep{vinyals2016matching}, Baseline++~\citep{chen2018closer}, MetaOptNet~\citep{lee2019meta}, and Meta-Baseline~\citep{chen2020new}, who only assume the input is a vector. %

\xhdr{Results} Table~\ref{tab:fewshot} shows that our \ours{} outperforms the standard few-shot learning benchmarks. Note that if we set KG Layer = 0, the proposed \ours{} degenerates to Meta-Baseline. The experiments clearly demonstrate the benefits of building a knowledge graph on the last layer. Furthermore, as an ablation study we manipulate the number of KG layers and find that in the few-shot image recognition setting, there's an improvement for stacking 2 KG layers instead of 1, meaning non-linearity is useful. We do not observe further improvements for more than 3 layers. 

\begin{table}[t]
\centering
\fontsize{9}{8}\selectfont
\caption{Results on 5-way, 5-shot classification on mini-ImageNet and tiered-ImageNet datasets. We report
the average accuracy and standard deviation over 800 randomly sampled episodes. \ours demonstrates consistent improvement from stacking KG layer = 0 to 2.
}
\label{tab:fewshot}
\begin{tabular}{c|cc}
\toprule
\texttt{Method}            & \texttt{mini-ImageNet}    & \texttt{tiered-ImageNet}     \\  \midrule
MatchNet~\citep{vinyals2016matching} &  78.72 $\pm$ 0.15 &  80.60 $\pm$ 0.71 \\
Baseline++~\citep{chen2018closer} & 77.76 $\pm$ 0.17 & 83.74 $\pm$ 0.18 \\
MetaOptNet~\citep{lee2019meta} & 78.63 $\pm$ 0.46 & 81.56 $\pm$ 0.53 \\ 
Meta-Baseline~\citep{chen2020new} \textsubscript{(KG layer = 0) } &  79.26 $ \pm $ 0.17 & 83.29 $ \pm $ 0.18 \\ \midrule
\textbf{\ours{} \textsubscript{(KG layer = 1)}} &  79.86 $ \pm $ 0.18 & 83.91 $ \pm $ 0.17 \\ 
\textbf{\ours{} \textsubscript{(KG layer = 2)}} & \textbf{81.13} $\pm$ \textbf{0.17} & \textbf{84.68} $\pm$ \textbf{0.17} \\ 
\bottomrule
\end{tabular}
\end{table}
\section{Related Work}
\xhdr{Multi-task learning}
Multi-task learning is a learning paradigm that jointly optimizes a set of tasks with shared parameters. It is generally believed that relations across different tasks can improve the overall performance. Some works cast it as a multi-objective optimization problem and introduce multiple gradient-based methods to reduce negative transfer among tasks~\citep{fliege2016method,lin2019pareto}. Other works assign (adaptive) weights for different tasks using certain heuristics~\citep{kendall2018multi,chen2018closer}. Our empirical study is also closely related to multi-label learning, where the problem is usually decomposed into multiple binary classification tasks~\citep{tsoumakas2007multi}. There is a line of work learning to leverage the relationship among tasks~\citep{haller2021unsupervised,zamir2020robust}. \citet{wang2016cnn} utilized recurrent neural networks to transform labels into embedded label vectors to learn the correlation among labels. The most recent work is ML-GCN~\citep{chen2019multi}, which uses a GCN to map label graph into a set of inter-dependent classifiers. Although the major motivation of our work is also about leveraging correlation among tasks, our problem formulation is new and thus yields a novel algorithm.

\xhdr{Exploring graph structure for data and tasks} There are prior works exploring graph structure for data points or tasks, and graph structure was proven to be effective in some tasks. \citet{satorras2018few} explores graph neural representations over \emph{data points only} for few-shot learning. Besides, some works study how to transfer knowledge among tasks through constructing a graph over \emph{task nodes/classifiers only}~\citep{liu2019learning,chen2020knowledge}. Along this direction, instead of building a fully connected graph, recent works utilize auxiliary task structure/knowledge graph to build the graph~\citep{chen2019multi,lee2018multi}. In contrast, and orthogonal to the papers above, we focus on modeling data-task relationships with the reinterpretation of the last layer. In addition, with data-task relationships solely, we are still able to capture data-data, task-task relationships implicitly through higher-order message passing.

\section{Conclusion}

We introduced relational multi-task settings in which the methods are required to learn to leverage labels on auxiliary tasks to predict on the new task. These settings are impactful in the biomedical domain where labels of different tasks are often scarcely available. To address these settings, we propose \ours, which is general enough to allow us formulating the above settings in a single framework. We demonstrated that \ours can successfully utilize the relations among tasks, outperforming the state-of-the-art methods under the proposed relational multi-task learning setting, with up to 27\% improvement in ROC AUC. We limit our focus to model data-task relationships since most benchmark datasets do not have information on data-data or task-task relationships, though \ours is expressive enough to model such relationships. We leave extending \ours{} to more complex relationships or tasks for future work.

\section*{Acknowledgement}

We want to give special thanks to Hamed Nilforoshan, Michihiro Yasunaga, Weihua Hu, Paridhi Maheshwari, Xuechen Li, Wengong Jin, Octavian Ganea, Tommi Jaakkola, and Regina Barzilay for the thoughtful discussions.
We gratefully acknowledge the support of DARPA under Nos. HR00112190039 (TAMI), N660011924033 (MCS); ARO under Nos. W911NF-16-1-0342 (MURI), W911NF-16-1-0171 (DURIP);
NSF under Nos. OAC-1835598 (CINES), OAC-1934578 (HDR), CCF-1918940 (Expeditions), IIS-2030477 (RAPID), NIH under No. R56LM013365;
Stanford Data Science Initiative, Wu Tsai Neurosciences Institute, Chan Zuckerberg Biohub,
Amazon, JPMorgan Chase, Docomo, Hitachi, Intel, KDDI, Toshiba, NEC, and UnitedHealth Group.
Jiaxuan You is supported by JPMC PhD Fellowship and Baidu Scholarship.

\bibliography{main}

\begin{thebibliography}{40}
\providecommand{\natexlab}[1]{#1}
\providecommand{\url}[1]{\texttt{#1}}
\expandafter\ifx\csname urlstyle\endcsname\relax
  \providecommand{\doi}[1]{doi: #1}\else
  \providecommand{\doi}{doi: \begingroup \urlstyle{rm}\Url}\fi

\bibitem[Cao et~al.(2020{\natexlab{a}})Cao, Brbic, and
  Leskovec]{cao2020concept}
Kaidi Cao, Maria Brbic, and Jure Leskovec.
\newblock Concept learners for few-shot learning.
\newblock In \emph{International Conference on Learning Representations},
  2020{\natexlab{a}}.

\bibitem[Cao et~al.(2020{\natexlab{b}})Cao, Ji, Cao, Chang, and
  Niebles]{cao2020few}
Kaidi Cao, Jingwei Ji, Zhangjie Cao, Chien-Yi Chang, and Juan~Carlos Niebles.
\newblock Few-shot video classification via temporal alignment.
\newblock In \emph{Proceedings of the IEEE/CVF Conference on Computer Vision
  and Pattern Recognition}, pp.\  10618--10627, 2020{\natexlab{b}}.

\bibitem[Caruana(1997)]{caruana1997multitask}
Rich Caruana.
\newblock Multitask learning.
\newblock \emph{Machine learning}, 28\penalty0 (1):\penalty0 41--75, 1997.

\bibitem[Chen et~al.(2020{\natexlab{a}})Chen, Chen, Hui, Wu, Li, and
  Lin]{chen2020knowledge}
Riquan Chen, Tianshui Chen, Xiaolu Hui, Hefeng Wu, Guanbin Li, and Liang Lin.
\newblock Knowledge graph transfer network for few-shot recognition.
\newblock In \emph{Proceedings of the AAAI Conference on Artificial
  Intelligence}, volume~34, pp.\  10575--10582, 2020{\natexlab{a}}.

\bibitem[Chen et~al.(2018)Chen, Liu, Kira, Wang, and Huang]{chen2018closer}
Wei-Yu Chen, Yen-Cheng Liu, Zsolt Kira, Yu-Chiang~Frank Wang, and Jia-Bin
  Huang.
\newblock A closer look at few-shot classification.
\newblock In \emph{International Conference on Learning Representations}, 2018.

\bibitem[Chen et~al.(2020{\natexlab{b}})Chen, Wang, Liu, Xu, and
  Darrell]{chen2020new}
Yinbo Chen, Xiaolong Wang, Zhuang Liu, Huijuan Xu, and Trevor Darrell.
\newblock A new meta-baseline for few-shot learning.
\newblock \emph{arXiv preprint arXiv:2003.04390}, 2020{\natexlab{b}}.

\bibitem[Chen et~al.(2019)Chen, Wei, Wang, and Guo]{chen2019multi}
Zhao-Min Chen, Xiu-Shen Wei, Peng Wang, and Yanwen Guo.
\newblock Multi-label image recognition with graph convolutional networks.
\newblock In \emph{Proceedings of the IEEE/CVF Conference on Computer Vision
  and Pattern Recognition}, pp.\  5177--5186, 2019.

\bibitem[Dempster et~al.(1977)Dempster, Laird, and Rubin]{dempster1977maximum}
Arthur~P Dempster, Nan~M Laird, and Donald~B Rubin.
\newblock Maximum likelihood from incomplete data via the em algorithm.
\newblock \emph{Journal of the Royal Statistical Society: Series B
  (Methodological)}, 39\penalty0 (1):\penalty0 1--22, 1977.

\bibitem[Finn et~al.(2017)Finn, Abbeel, and Levine]{finn2017model}
Chelsea Finn, Pieter Abbeel, and Sergey Levine.
\newblock Model-agnostic meta-learning for fast adaptation of deep networks.
\newblock In \emph{International Conference on Machine Learning}, pp.\
  1126--1135. PMLR, 2017.

\bibitem[Fliege \& Vaz(2016)Fliege and Vaz]{fliege2016method}
Jorg Fliege and A~Ismael~F Vaz.
\newblock A method for constrained multiobjective optimization based on sqp
  techniques.
\newblock \emph{SIAM Journal on Optimization}, 26\penalty0 (4):\penalty0
  2091--2119, 2016.

\bibitem[Gilmer et~al.(2017)Gilmer, Schoenholz, Riley, Vinyals, and
  Dahl]{gilmer2017neural}
Justin Gilmer, Samuel~S Schoenholz, Patrick~F Riley, Oriol Vinyals, and
  George~E Dahl.
\newblock Neural message passing for quantum chemistry.
\newblock In \emph{International Conference on Machine Learning (ICML)}, 2017.

\bibitem[Haller et~al.(2021)Haller, Burceanu, and
  Leordeanu]{haller2021unsupervised}
Emanuela Haller, Elena Burceanu, and Marius Leordeanu.
\newblock Unsupervised domain adaptation through iterative consensus shift in a
  multi-task graph.
\newblock \emph{arXiv preprint arXiv:2103.14417}, 2021.

\bibitem[Hamilton et~al.(2017)Hamilton, Ying, and
  Leskovec]{hamilton2017inductive}
William~L Hamilton, Rex Ying, and Jure Leskovec.
\newblock Inductive representation learning on large graphs.
\newblock In \emph{Proceedings of the 31st International Conference on Neural
  Information Processing Systems}, pp.\  1025--1035, 2017.

\bibitem[He et~al.(2019)He, Xie, Rong, Huang, Li, Huang, Ren, and
  Shahabi]{he2019bipartite}
Chaoyang He, Tian Xie, Yu~Rong, Wenbing Huang, Yanfang Li, Junzhou Huang, Xiang
  Ren, and Cyrus Shahabi.
\newblock Bipartite graph neural networks for efficient node representation
  learning.
\newblock \emph{arXiv e-prints}, pp.\  arXiv--1906, 2019.

\bibitem[Huang et~al.(2016)Huang, Xia, Nguyen, Zhao, Sakamuru, Zhao, Shahane,
  Rossoshek, and Simeonov]{huang2016tox21challenge}
Ruili Huang, Menghang Xia, Dac-Trung Nguyen, Tongan Zhao, Srilatha Sakamuru,
  Jinghua Zhao, Sampada~A Shahane, Anna Rossoshek, and Anton Simeonov.
\newblock Tox21challenge to build predictive models of nuclear receptor and
  stress response pathways as mediated by exposure to environmental chemicals
  and drugs.
\newblock \emph{Frontiers in Environmental Science}, 3:\penalty0 85, 2016.

\bibitem[Kendall et~al.(2018)Kendall, Gal, and Cipolla]{kendall2018multi}
Alex Kendall, Yarin Gal, and Roberto Cipolla.
\newblock Multi-task learning using uncertainty to weigh losses for scene
  geometry and semantics.
\newblock In \emph{Proceedings of the IEEE conference on computer vision and
  pattern recognition}, pp.\  7482--7491, 2018.

\bibitem[Koller \& Friedman(2009)Koller and Friedman]{koller2009probabilistic}
Daphne Koller and Nir Friedman.
\newblock \emph{Probabilistic graphical models: principles and techniques}.
\newblock MIT press, 2009.

\bibitem[Kuhn et~al.(2016)Kuhn, Letunic, Jensen, and Bork]{kuhn2016sider}
Michael Kuhn, Ivica Letunic, Lars~Juhl Jensen, and Peer Bork.
\newblock The sider database of drugs and side effects.
\newblock \emph{Nucleic acids research}, 44\penalty0 (D1):\penalty0
  D1075--D1079, 2016.

\bibitem[Lee et~al.(2018)Lee, Fang, Yeh, and Wang]{lee2018multi}
Chung-Wei Lee, Wei Fang, Chih-Kuan Yeh, and Yu-Chiang~Frank Wang.
\newblock Multi-label zero-shot learning with structured knowledge graphs.
\newblock In \emph{IEEE Computer Society Conference on Computer Vision and
  Pattern Recognition (CVPR)}, 2018.

\bibitem[Lee et~al.(2019)Lee, Maji, Ravichandran, and Soatto]{lee2019meta}
Kwonjoon Lee, Subhransu Maji, Avinash Ravichandran, and Stefano Soatto.
\newblock Meta-learning with differentiable convex optimization.
\newblock In \emph{Proceedings of the IEEE/CVF Conference on Computer Vision
  and Pattern Recognition}, pp.\  10657--10665, 2019.

\bibitem[Lin et~al.(2014)Lin, Maire, Belongie, Hays, Perona, Ramanan,
  Doll{\'a}r, and Zitnick]{lin2014microsoft}
Tsung-Yi Lin, Michael Maire, Serge Belongie, James Hays, Pietro Perona, Deva
  Ramanan, Piotr Doll{\'a}r, and C~Lawrence Zitnick.
\newblock Microsoft coco: Common objects in context.
\newblock In \emph{European conference on computer vision}, pp.\  740--755.
  Springer, 2014.

\bibitem[Lin et~al.(2019)Lin, Zhen, Li, Zhang, and Kwong]{lin2019pareto}
Xi~Lin, Hui-Ling Zhen, Zhenhua Li, Qingfu Zhang, and Sam Kwong.
\newblock Pareto multi-task learning.
\newblock In \emph{NIPS}, 2019.

\bibitem[Liu et~al.(2019)Liu, Zhou, Long, Jiang, and Zhang]{liu2019learning}
Lu~Liu, Tianyi Zhou, Guodong Long, Jing Jiang, and Chengqi Zhang.
\newblock Learning to propagate for graph meta-learning.
\newblock 2019.

\bibitem[Lu et~al.(2019)Lu, Liu, Wang, Huang, Lin, and He]{lu2019molecular}
Chengqiang Lu, Qi~Liu, Chao Wang, Zhenya Huang, Peize Lin, and Lixin He.
\newblock Molecular property prediction: A multilevel quantum interactions
  modeling perspective.
\newblock In \emph{AAAI Conference on Artificial Intelligence (AAAI)}, 2019.

\bibitem[Ma et~al.(2020)Ma, Du, and Matusik]{ma2020efficient}
Pingchuan Ma, Tao Du, and Wojciech Matusik.
\newblock Efficient continuous pareto exploration in multi-task learning.
\newblock In \emph{International Conference on Machine Learning}, pp.\
  6522--6531. PMLR, 2020.

\bibitem[Paszke et~al.(2017)Paszke, Gross, Chintala, Chanan, Yang, DeVito, Lin,
  Desmaison, Antiga, and Lerer]{paszke2017automatic}
Adam Paszke, Sam Gross, Soumith Chintala, Gregory Chanan, Edward Yang, Zachary
  DeVito, Zeming Lin, Alban Desmaison, Luca Antiga, and Adam Lerer.
\newblock Automatic differentiation in pytorch.
\newblock 2017.

\bibitem[Ren et~al.(2018)Ren, Triantafillou, Ravi, Snell, Swersky, Tenenbaum,
  Larochelle, and Zemel]{ren2018meta}
Mengye Ren, Eleni Triantafillou, Sachin Ravi, Jake Snell, Kevin Swersky,
  Joshua~B Tenenbaum, Hugo Larochelle, and Richard~S Zemel.
\newblock Meta-learning for semi-supervised few-shot classification.
\newblock In \emph{International Conference on Learning Representations}, 2018.

\bibitem[Richard et~al.(2016)Richard, Judson, Houck, Grulke, Volarath,
  Thillainadarajah, Yang, Rathman, Martin, Wambaugh,
  et~al.]{richard2016toxcast}
Ann~M Richard, Richard~S Judson, Keith~A Houck, Christopher~M Grulke, Patra
  Volarath, Inthirany Thillainadarajah, Chihae Yang, James Rathman, Matthew~T
  Martin, John~F Wambaugh, et~al.
\newblock Toxcast chemical landscape: paving the road to 21st century
  toxicology.
\newblock \emph{Chemical research in toxicology}, 29\penalty0 (8):\penalty0
  1225--1251, 2016.

\bibitem[Rong et~al.(2020)Rong, Bian, Xu, Xie, Wei, Huang, and
  Huang]{rong2020self}
Yu~Rong, Yatao Bian, Tingyang Xu, Weiyang Xie, Ying Wei, Wenbing Huang, and
  Junzhou Huang.
\newblock Self-supervised graph transformer on large-scale molecular data.
\newblock \emph{Advances in Neural Information Processing Systems (NeurIPS)},
  2020.

\bibitem[Satorras \& Estrach(2018)Satorras and Estrach]{satorras2018few}
Victor~Garcia Satorras and Joan~Bruna Estrach.
\newblock Few-shot learning with graph neural networks.
\newblock In \emph{International Conference on Learning Representations}, 2018.

\bibitem[Sener \& Koltun(2018)Sener and Koltun]{senermulti}
Ozan Sener and Vladlen Koltun.
\newblock Multi-task learning as multi-objective optimization.
\newblock In \emph{NIPS}, 2018.

\bibitem[Tsoumakas \& Katakis(2007)Tsoumakas and Katakis]{tsoumakas2007multi}
Grigorios Tsoumakas and Ioannis Katakis.
\newblock Multi-label classification: An overview.
\newblock \emph{International Journal of Data Warehousing and Mining (IJDWM)},
  3\penalty0 (3):\penalty0 1--13, 2007.

\bibitem[Turhan \& Bilge(2018)Turhan and Bilge]{turhan2018recent}
Ceren~G{\"u}zel Turhan and Hasan~Sakir Bilge.
\newblock Recent trends in deep generative models: a review.
\newblock In \emph{2018 3rd International Conference on Computer Science and
  Engineering (UBMK)}, pp.\  574--579. IEEE, 2018.

\bibitem[Vinyals et~al.(2016)Vinyals, Blundell, Lillicrap, Kavukcuoglu, and
  Wierstra]{vinyals2016matching}
Oriol Vinyals, Charles Blundell, Timothy Lillicrap, Koray Kavukcuoglu, and Daan
  Wierstra.
\newblock Matching networks for one shot learning.
\newblock In \emph{Proceedings of the 30th International Conference on Neural
  Information Processing Systems}, pp.\  3637--3645, 2016.

\bibitem[Wang et~al.(2016)Wang, Yang, Mao, Huang, Huang, and Xu]{wang2016cnn}
Jiang Wang, Yi~Yang, Junhua Mao, Zhiheng Huang, Chang Huang, and Wei Xu.
\newblock Cnn-rnn: A unified framework for multi-label image classification.
\newblock In \emph{Proceedings of the IEEE conference on computer vision and
  pattern recognition}, pp.\  2285--2294, 2016.

\bibitem[Xiong et~al.(2019)Xiong, Wang, Liu, Zhong, Wan, Li, Li, Luo, Chen,
  Jiang, et~al.]{xiong2019pushing}
Zhaoping Xiong, Dingyan Wang, Xiaohong Liu, Feisheng Zhong, Xiaozhe Wan, Xutong
  Li, Zhaojun Li, Xiaomin Luo, Kaixian Chen, Hualiang Jiang, et~al.
\newblock Pushing the boundaries of molecular representation for drug discovery
  with the graph attention mechanism.
\newblock \emph{Journal of medicinal chemistry}, 2019.

\bibitem[Xue et~al.(2021)Xue, Yang, Rajan, Jiang, Wei, and
  Lin]{xue2021multiplex}
Hansheng Xue, Luwei Yang, Vaibhav Rajan, Wen Jiang, Yi~Wei, and Yu~Lin.
\newblock Multiplex bipartite network embedding using dual hypergraph
  convolutional networks.
\newblock In \emph{Proceedings of the Web Conference 2021}, pp.\  1649--1660,
  2021.

\bibitem[Yang et~al.(2019)Yang, Swanson, Jin, Coley, Eiden, Gao, Guzman-Perez,
  Hopper, Kelley, Mathea, et~al.]{yang2019analyzing}
Kevin Yang, Kyle Swanson, Wengong Jin, Connor Coley, Philipp Eiden, Hua Gao,
  Angel Guzman-Perez, Timothy Hopper, Brian Kelley, Miriam Mathea, et~al.
\newblock Analyzing learned molecular representations for property prediction.
\newblock \emph{Journal of chemical information and modeling}, 2019.

\bibitem[Zamir et~al.(2018)Zamir, Sax, Shen, Guibas, Malik, and
  Savarese]{zamir2018taskonomy}
Amir~R Zamir, Alexander Sax, William Shen, Leonidas~J Guibas, Jitendra Malik,
  and Silvio Savarese.
\newblock Taskonomy: Disentangling task transfer learning.
\newblock In \emph{Proceedings of the IEEE conference on computer vision and
  pattern recognition}, pp.\  3712--3722, 2018.

\bibitem[Zamir et~al.(2020)Zamir, Sax, Cheerla, Suri, Cao, Malik, and
  Guibas]{zamir2020robust}
Amir~R Zamir, Alexander Sax, Nikhil Cheerla, Rohan Suri, Zhangjie Cao, Jitendra
  Malik, and Leonidas~J Guibas.
\newblock Robust learning through cross-task consistency.
\newblock In \emph{Proceedings of the IEEE/CVF Conference on Computer Vision
  and Pattern Recognition}, pp.\  11197--11206, 2020.

\end{thebibliography}
\bibliographystyle{iclr2022_conference}

\newpage
\appendix

\section{Additional Implementation Details}
\label{sec:implementation}

\xhdr{Implementation details for biomedical datasets}
We use an adapted version of GraphSAGE \cite{hamilton2017inductive} as the base neural architecture, and implement MetaLink based on the descriptions in the main paper. We use Adam optimizer, with initial learning of $0.001$ and cosine learning rate scheduler. The model is trained with a batch size of 128 for 50 epochs. We search over the number of layers of $[2,3,4,5]$, and report the test set performance when the best validation set performance is reached.

\xhdr{Implementation details for MS-COCO}
We use ResNet-50 with an input size of 224$\times$224 as the base neural architecture, and implement MetaLink based on the descriptions in the main paper. We perform standard data augmentation, including random crop and horizontal flipping. We use Adam optimizer with an initial learning rate of $0.002$, weight decay of $1\times10^{-4}$. The model is trained with a batch size of 256 for 40 epochs. We use the same set of hyper-parameters for all the four settings mentioned in the main paper.

\xhdr{Implementation details for few-shot learning}
We use the widely adopted backbones ResNet-12 with an input size of 84$\times$84~\citep{chen2020new}. We perform standard data
augmentation, including random crop and horizontal flipping. We use SGD optimizer with an initial learning rate of $0.001$, momentum of 0.9, weight decay of $5\times10^{-4}$. The model is trained with a batch size of 4 for 20 epochs. Notice one batch includes a support set (with 5$\times$5 examples) and a query set(with 15 examples). In a nutshell, we use the same set of hyperparameters as~\citep{chen2020new}, which is exactly the degenerated model if we set KG layer = 0.

\xhdr{Additional algorithms}
We summarize the steps of \ours{} for relational setting in Algorithm~\ref{alg:relational}, for meta setting in Algorithm~\ref{alg:meta}.

\xhdr{Code release}
We will make all the source code public at the time of publication.

\begin{algorithm}[h]
\caption{\ours{} Training in Relational Setting}
\label{alg:relational}
\begin{algorithmic}[1]
\Require Dataset $\mathcal{D}_\text{train} = \{(\mb{x},y)\}$. A parameterized embedding function $f_\theta$. Last layer weights for each task $
\{ \mb{w}_j \}$. A parameterized heterogeneous GNN $f_\phi$. Number of GNN layers $L$. 

\For {each iteration }
    
    \State $ \{ (\mb{x},  \{  y_{j \in T^{(i)}_{\text{aux}}}^{(i)} \} , \{  y_{j \in T^{(i)}_{\text{test}}}^{(i)} \}  \} \leftarrow \text{SampleMiniBatch}(\mathcal{D}_\text{train}) $
    \State $ \{ \mb{z} \}  \leftarrow f_\theta ( \mb{x} ) \text{ for } \mb{x} \in \{ (\mb{x},  \{  y_{j \in T^{(i)}_{\text{aux}}}^{(i)} \}  , \{  y_{j \in T^{(i)}_{\text{test}}}^{(i)} \}  \}$
    \State $V_d^{(0)} = \{ \mb{h}_i^{(0)} \leftarrow \mb{z} \text{ for } \mb{z} \in \{ \mb{z} \} \}$ \Comment{Initialize data nodes}
    \State $V_t^{(0)} = \{ \mb{h}_j^{(0)} \leftarrow \mb{w}_j \text{ for } \mb{w}_j \in \{ \mb{w}_j \} $ \Comment{Initialize task nodes}
    \State $E = \{ \mb{e}_{ij} \leftarrow (\mb{x}^{(i)}, t_j) \text{ for }  y^{(i)}_j \in \{  y_{j \in T_{\text{aux}}}^{(i)} \} $ \Comment{Initialize edges}
    
    \For {$l=1$ to $L$ }
        \State $ V_d^{(l)}, V_t^{(l)} \leftarrow  \text{GraphConv} (V_d^{(l-1)}, V_t^{(l-1)}, E)$ with $f_\phi$
    \EndFor
    \State $\text{logits} \leftarrow \text{EdgePred}( V_d^{(L)}, V_t^{(L)})$ with $f_\phi$
    \State $\text{Backward}\left( \text{Criterion}(\text{logits}, \{  \{ y_j^{(i)}  \}_{j \in T^{(i)}_{\text{test}}} ) \right)$
\EndFor

\end{algorithmic}
\end{algorithm}

\begin{algorithm}[h]
\caption{\ours{} Training in Meta Setting}
\label{alg:meta}
\begin{algorithmic}[1]
\Require Dataset $\mathcal{D}_\text{train} = \{(\mb{x},y)\}$. A parameterized embedding function $f_\theta$. A parameterized heterogeneous GNN $f_\phi$. Number of GNN layers $L$. 

\For {each iteration }
    
    \State $S, Q \leftarrow \text{SampleMiniBatch}(\mathcal{D}_\text{train}) $ \Comment{Simulate meta setting in training}
    \State $ \{ \mb{z} \}  \leftarrow f_\theta ( \mb{x} ) \text{ for } \mb{x} \in (S, Q)$
    \State $V_d^{(0)} = \{ \mb{h}_i^{(0)} \leftarrow \mb{z} \text{ for } \mb{z} \in \{ \mb{z} \} \}$ \Comment{Initialize data nodes}
    \State $V_t^{(0)} = \{ \mb{h}_j^{(0)} \leftarrow \textbf{1}\}$ \Comment{Initialize task nodes}
    \State $E = \{ \mb{e}_{ij} \leftarrow (\mb{x}^{(i)}, t_j) \text{ for }  y^{(i)}_j \in S \} $ \Comment{Initialize edges}
    
    \For {$l=1$ to $L$ }
        \State $ V_d^{(l)}, V_t^{(l)} \leftarrow  \text{GraphConv} (V_d^{(l-1)}, V_t^{(l-1)}, E)$ with $f_\phi$
    \EndFor
    \State $\text{logits} \leftarrow \text{EdgePred}( V_d^{(L)}, V_t^{(L)})$ with $f_\phi$
    \State $\text{Backward}\left( \text{Criterion}(\text{logits}, \{  \{ y_j^{(i)}  \}_{j \in T_{\text{s}}}  \} \in Q)\right)$
\EndFor

\end{algorithmic}
\end{algorithm}

\section{Additional Discussions}
\label{sec:discussions}

\xhdr{Limitations}
i) In principle, \ours leverages task correlation to achieve data efficiency/gain better performance. Thus, it is less effective when tasks are not correlated. We believe it is fair to assume a multi-task learning system will work worse when tasks as not correlated. ii) Though our method is general to model diverse multi-task scenarios, we limit our empirical study to multi-label learning mostly due to the limitation of dataset availability. We hope that the proposed \ours can inspire researchers to collect more multi-task learning datasets with diverse use cases.

\xhdr{Social impact} 
In practice, collecting high-quality datasets is often expensive and
time-consuming. In the biomedical domain, labeling requires domain expertise and hence
is very resource-intensive. In some applications, it requires long-term experiments to get the ground truth labels. The approach that we proposed could reduce the high labeling cost by utilizing auxiliary task labels that are already available. 
In principle, \ours{} attempts to leverage correlation among tasks. Thus, when it comes to future deployment, we suggest being careful about defining tasks. Because it might come across similar issues related to fairness as in other supervised learning problems, where the algorithm learns biased correlations from the datasets that are not considered to be appropriate.
The datasets we used for experiments are among the most widely-used benchmarks, which should not contain any undesirable bias.

\section{Additional Results}
\label{sec:results}

\xhdr{Advantages on using heterogeneous weights}
As discussed in Section 3, we proposed to learn from the knowledge graph constructed via a heterogeneous GNN. Here we empirically demonstrate the improvement of using distinct weights for updating data nodes and task nodes. Results are summarized in Table~\ref{tab:hetero}.

\begin{table}[h]
\caption{Results of ROC AUC (for biomedical) and mAP (for MS-COCO) datasets.
}
\label{tab:hetero}
\centering
\begin{footnotesize}

\resizebox{\textwidth}{!}{
\begin{tabular}{cc|ccccc}
\toprule

\texttt{Setting} & \texttt{Weights} & \texttt{Tox21} (12 tasks) & \texttt{Sider} (27 tasks) & \texttt{ToxCast} (617 tasks) & \texttt{MS-COCO} (80 tasks) \\ \midrule
\multirow{2}{*}{Relational} & Shared & 83.1 $\pm$ 1.6 & 75.8 $\pm$ 2.5 & 78.1 $\pm$ 1.3 & 74.02 $\pm$ 0.16 \\ 
& Hetero & \textbf{83.7} $\pm$ \textbf{1.9}  & \textbf{76.8} $\pm$ \textbf{3.0} & \textbf{79.4} $\pm$ \textbf{1.0} & \textbf{75.36} $\pm$ \textbf{0.16}  \\

\bottomrule
\end{tabular}
}

\label{tab:graph_property}
\end{footnotesize}
\end{table}

\end{document}